\documentclass[runningheads]{llncs}

\usepackage{color}
\definecolor{green}{rgb}{0, 0.5, 0}
\definecolor{orange}{rgb}{0.8, 0.6, 0.2}
\definecolor{red}{rgb}{1.0, 0.0, 0.0}
\definecolor{teal}{rgb}{0.0, 0.4, 0.4}
\definecolor{purple}{rgb}{0.65,0,0.65}
\definecolor{saffron}{rgb}{0.95,0.75,0.2}
\definecolor{turquoise}{rgb}{0.0,0.5,0.5}
\definecolor {mygray}{gray}{.9}

\usepackage{booktabs}
\usepackage{colortbl}
\usepackage{makecell}
\usepackage{multirow}
\usepackage{multicol}
\usepackage{array}
\usepackage{gensymb}
\newcolumntype{L}[1]{>{\raggedright\arraybackslash}p{#1}}
\newcolumntype{C}[1]{>{\centering\arraybackslash}p{#1}}
\newcolumntype{R}[1]{>{\raggedleft\arraybackslash}p{#1}}
\usepackage{amsmath}

\usepackage{tabu}
\usepackage{graphicx}

\usepackage{eccv}
\usepackage{eccvabbrv}
\usepackage{graphicx}
\usepackage{booktabs}
\usepackage[accsupp]{axessibility}
\usepackage{ulem}
\usepackage{hyperref}

\definecolor{mzx}{rgb}{0.95, 0.1, 0.1}
\definecolor{ls}{rgb}{0.12, 0.6, 0.14}

\usepackage{orcidlink}

\begin{document}

\title{RoadPainter: Points Are Ideal Navigators for Topology transformER} 
\titlerunning{RoadPainter: Points Are Ideal Navigators for Topology transformER}
\author{Zhongxing Ma$^\ast$ \hspace{0.1cm} Shuang Liang$^\ast$ \hspace{0.1cm} Yongkun Wen$^\ast$ \hspace{0.1cm} Weixin Lu \hspace{0.1cm} Guowei Wan$^\dagger$}
\authorrunning{Z. Ma, S. Liang, Y. Wen, W. Lu, G. Wan}
\institute{Baidu Autonomous Driving Technology Department (ADT)\\
\email{\{mazhongxing, liangshuang18, wenyongkun, luweixin, wanguowei\}@baidu.com}
\def\thefootnote{$\ast$}\footnotetext{Authors with equal contributions}
\def\thefootnote{$\dagger$}\footnotetext{Author to whom correspondence should be addressed}
}

\maketitle

\begin{abstract}
Topology reasoning aims to provide a precise understanding of road scenes, enabling autonomous systems to identify safe and efficient routes.
In this paper, we present RoadPainter, an innovative approach for detecting and reasoning the topology of lane centerlines using multi-view images.
The core concept behind RoadPainter is to extract a set of points from each centerline mask to improve the accuracy of centerline prediction.
We start by implementing a transformer decoder that integrates a hybrid attention mechanism and a real-virtual separation strategy to predict coarse lane centerlines and establish topological associations.
Then, we generate centerline instance masks guided by the centerline points from the transformer decoder.
Moreover, we derive an additional set of points from each mask and combine them with previously detected centerline points for further refinement.
Additionally, we introduce an optional module that incorporates a Standard Definition (SD) map to further optimize centerline detection and enhance topological reasoning performance.
Experimental evaluations on the OpenLane-V2 dataset demonstrate the state-of-the-art performance of RoadPainter.
\end{abstract}
\section{Introduction}
\label{section:intro}

In recent years, the field of topology reasoning in autonomous driving or advanced assisted driving has gained increasing attention. 
The primary objective of this research area is to extract lane centerlines and their topological relationships from onboard sensor data. 
This extraction process plays a crucial role in providing accurate and comprehensive online routing information, benefiting downstream tasks such as trajectory prediction\cite{mtr++, multipath++, hivt, wayformer} and planning\cite{uniad, p3, st-p3}.
Traditional approaches to constructing lane centerlines often frame the task as a map elements detection task. 
Some approaches\cite{LaneNet,UFAST,CondLaneNet,GANet} detect lanelines in image space and then use image-to-ground projection to convert points into 3D space for downstream use. 
Despite exhibiting remarkable accuracy in simple scenes, these lane detection algorithms face limitations in complex scenarios due to the planar assumption and pitch error in image-to-ground projection.

Recent advancements in the field\cite{VectorMapNet, MapTR, BeMapNet} have showcased progress in extracting lanelines from multi-view images and representing them in the unified Bird's Eye View (BEV) space. 
Approaches like VectorMapNet\cite{VectorMapNet} and MapTR\cite{MapTR} have succeeded in constructing online vectorized maps on BEV space without the need for post-processing laneline reprojection.
However, these methods mainly focus on extracting distinct lane boundaries, without capturing the representations of lane instances and their topological connections.
STSU\cite{STSU} introduces a method that detects lane instances and predicts lane topological relationships in the BEV space by describing each lane instance with a centerline. 
Subsequently, similar methods such as\cite{TPLR, LaneGAP} were proposed. 
TopoNet\cite{TopoNet} further enhances this task by not only predicting centerlines but also traffic elements. 
However, these methods predict centerline points via direct regression, which may be insufficient in challenging scenarios like areas with high curvature.

\begin{figure}[!!t]
	\centering
	\includegraphics[width=1.0\linewidth]{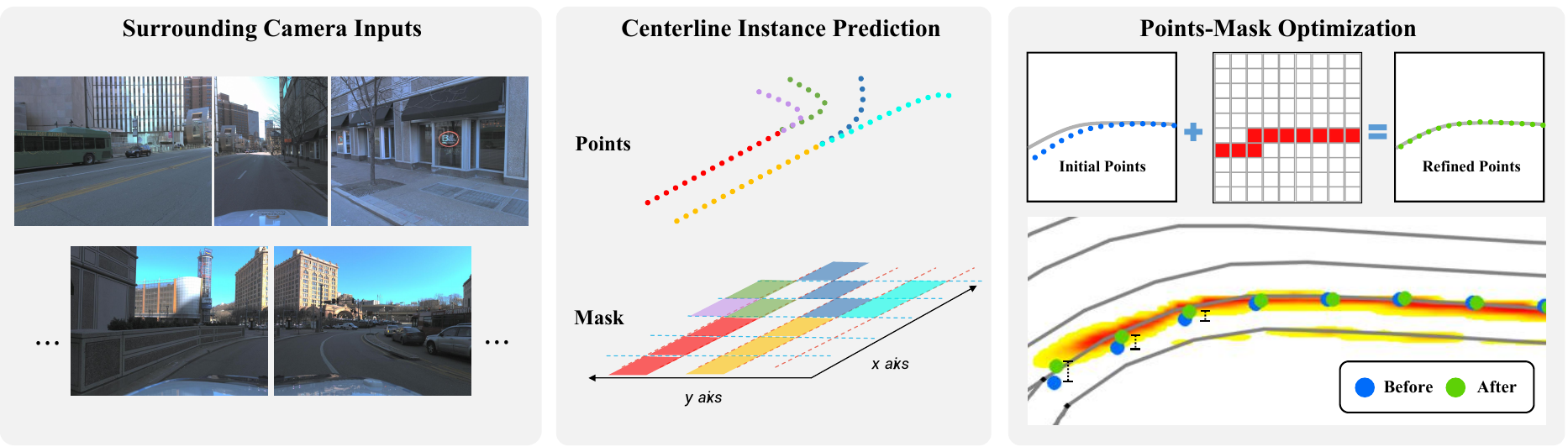}
	\caption{
		The illustration of our method.
		(a) Our model takes multi-view images as input.
		(b) From the constructed BEV features, two types of lane centerlines are derived: points and mask.
		(c) The initially detected points provide a rough localization, while the mask points enhance geometric details by utilizing the corresponding fine-grained heatmap.
	}
	\label{fig:teaser}
\end{figure}

To this end, we introduce a novel framework, RoadPainter, that initially regresses a set of centerline points for each lane and then refines these points using centerline instance masks. 
The word “Painter” is an acronym derived from the phrase “Points Are Ideal Navigators for Topology transformER”.
Regression provides stable initial localization but tends to favor straight-line data, causing methods like TopoNet to primarily learn straight centerlines. 
Segmentation offers accurate geometric details, improving the overall geometric accuracy of the centerlines.
Hence, our aim is to amalgamate the advantages of both techniques to optimize performance.
As illustrated in Fig.~\ref{fig:teaser}, our method utilizes surrounding images to construct BEV features, regresses centerline points and reasons topology relationships with a transformer decoder, and generates centerline instance masks guided by the regressed points.
A points-mask fusion module further refines the regressed centerline points with the aid of centerline instance masks. 
Experimental results on the OpenLane-V2 dataset\cite{OpenLaneV2} demonstrate the superior performance of RoadPainter, outperforming the state-of-the-art TopoNet by 7.7\% on the $\text{DET}_l$ metric. 
Furthermore, by leveraging the SD map, our proposed method exhibits superior performance compared to the results reported in\cite{SMERF}.

In summary, our main contributions are as follows:
\begin{itemize}
	\item Introduction of RoadPainter, a novel framework for centerline instance prediction and topological reasoning with refinement from centerline masks, ensuring the accurate preservation of geometric shapes of centerlines.
	\item Proposal of a novel module for generating centerline masks by incorporating initial regressed points, followed by the refinement of regressed points using masks in an end-to-end way, eliminating the necessity for post-processing.
	\item Demonstration of the effectiveness of our method through comprehensive tests and detailed ablations, showcasing state-of-the-art performance in lane detection and topology reasoning on the OpenLane-V2 dataset\cite{OpenLaneV2}.
\end{itemize}

\section{Related Work}
\label{section:related}

\subsection{Topology Reasoning}
\label{subsec:topology}

The first approach to understanding road structure through lane topology, as described in the study by STSU~\cite{STSU}, focused primarily on centerline detection.
The method establishes a three-stage paradigm consisting of BEV feature construction, centerline detection, and connection. 
This paradigm has been widely adopted in most topology reasoning papers.
Subsequently, TPLR~\cite{TPLR} enhanced the performance by incorporating the concept of the minimum circle formed by the directed centerline segment.
The path-wise representation of the centerline, as proposed in LaneGAP~\cite{LaneGAP}, effectively maintains the continuity and shape accuracy of the centerline. 
In order to ensure compliance with traffic regulations at intersections, TopoNet~\cite{TopoNet} employs Graph Neural Networks (GNN) to establish meaningful connections between driving lanes and traffic signs. 
By incorporating spatial position encoding, TopoMLP~\cite{TopoMLP} improves the performance of lane relation topology.
To further enhance the centerline detection accuracy, the study conducted by SMERF~\cite{SMERF} investigates the utilization of SD map.
Insights from 3D object detection, such as leveraging the relationship between objects and the driving lane, can enhance the quality of lane centerline extraction, as effectively demonstrated by the OLC method~\cite{OLC}.

\subsection{Online Map Construction}
\label{subsec:onlinemap}

The field of online map construction is highly relevant to lane topological reasoning.
Recent works aim to construct local High Definition (HD) map within a predefined range from onboard sensors, incorporating map elements like lane lines, pedestrian crossings, and curbs.
HDMapNet~\cite{HDMapNet} employs semantic learning on BEV features to extract map elements, which are then vectorized through post-processing.
Recent advancements in end-to-end learning are being used to eliminate post-processing steps. 
VectorMapNet~\cite{VectorMapNet} utilizes a methodology that identifies coarse boxes of map elements and then generates fine-grained geometric details.
InstaGram~\cite{InstaGram} employs a unique approach that involves initially identifying the geometric vertices of map elements, and then leveraging GNN to learn the associations between these vertices.
MapTR~\cite{MapTR} and its improved version ~\cite{MapTRv2} decouple the dependency between point association and positioning by assigning a fixed number of points to each map element. 
An alternative approach called BeMapNet~\cite{BeMapNet} utilizes a unified piecewise Bezier curve method that dynamically adjusts the number of points within a certain range.
Pivotnet~\cite{Pivotnet} introduces dynamic matching, retaining only points with relevant geometric features of the map element. 
ScalableMap~\cite{ScalableMap} enhances long-range perceptual capability by leveraging BEV features guided by the linear structure of map elements.

\subsection{Lane Detection}
\label{subsec:lanedet}

We also include the subject of lane detection due to its similarities with the aforementioned topics.
The early learning-based approaches, including LaneNet~\cite{LaneNet}, UFAST~\cite{UFAST}, CondLaneNet~\cite{CondLaneNet}, and GANet~\cite{GANet}, extract lane lines from 2D images.
LaneNet~\cite{LaneNet} formulates lane detection as an instance segmentation problem and employs two embedding branches. 
UFAST~\cite{UFAST} adopts feature aggregation and constructs segmentation on a smaller feature map to enhance detection speed.
CondLaneNet~\cite{CondLaneNet} utilizes shape prior and predicts the line location for each row in the image space, aiming to resolve the instance-level discrimination problem. 
GANet~\cite{GANet} reformulates the lane detection problem by directly regressing each keypoint to the starting point of the lane line.
To achieve more precise lane line detection, researchers have also explored lane detection in 3D space. 
3D-LaneNet~\cite{3D-LaneNet} introduces an architecture that generates 3D lane lines using two pathways: the image-view pathway encodes features in 2D images, while the top-view pathway offers translation-invariant features for 3D lane detection. 
PersFormer~\cite{PersFormer} adopts a unified 2D and 3D lane detection framework to simultaneously detect 2D and 3D lane lines.
BEV-LaneDet~\cite{BEV-LaneDet} proposes the use of a virtual camera to unify the intrinsic and extrinsic camera parameters, thereby enhancing the performance.
\section{Method}
\label{section:method}

\begin{figure*}[!t]
	\centering
	\includegraphics[width=\linewidth]{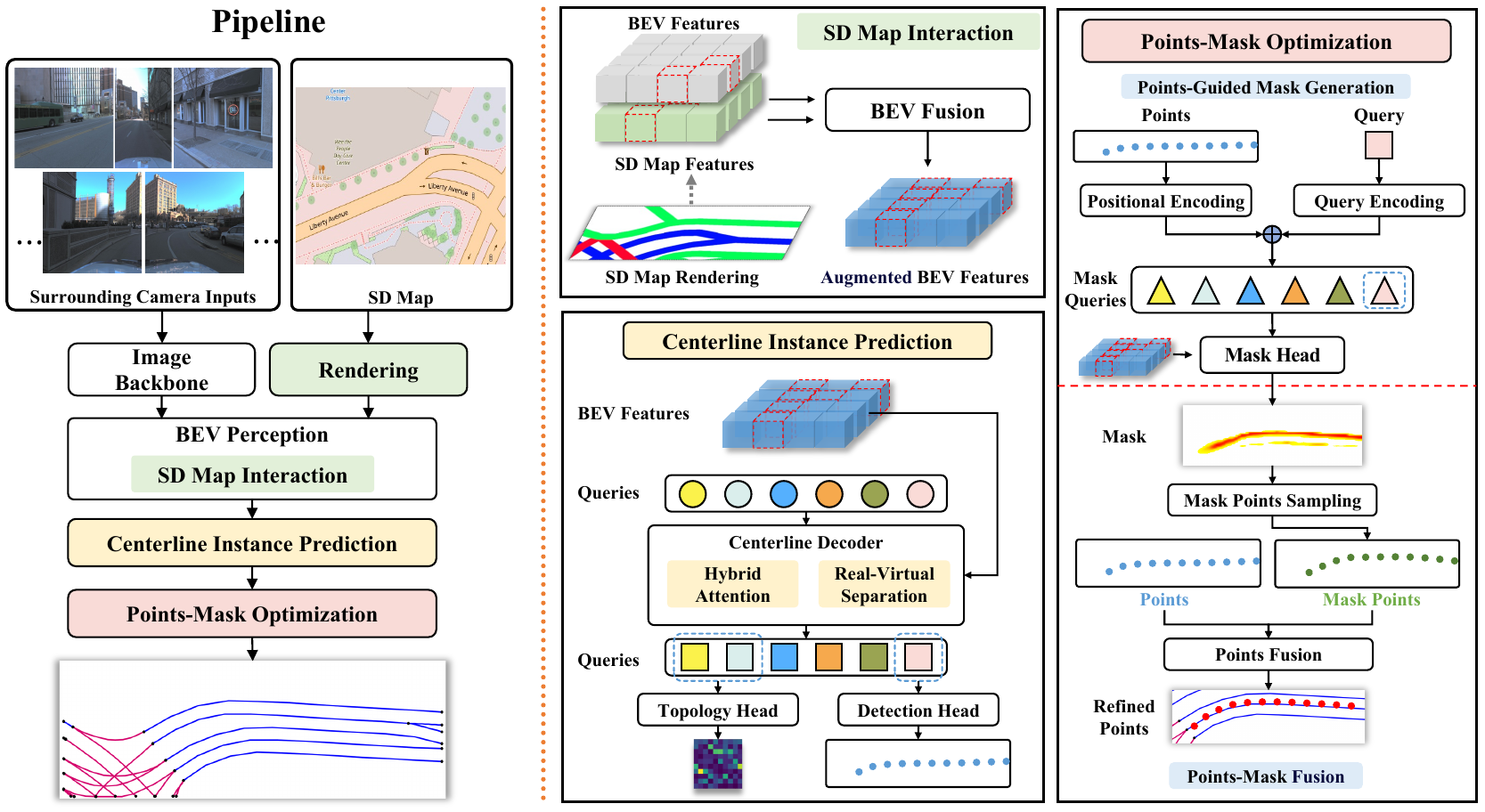}
	\caption{
		The architecture of RoadPainter consists of three main parts. 
		In the BEV perception stage, we convert sensor inputs and an optional SD Map into enhanced BEV features. 
		Next, we detect the lane centerlines by regressing centerline points and identifying lane-related topological associations. 
		Furthermore, we generate a mask for each centerline guided by the regressed centerline points.
		Lastly, we sample the mask and optimize them by incorporating the centerline points to improve the accuracy of centerlines and the reliability of topology. 
	}
	\label{fig:network}
\end{figure*}

Our method takes surrounding images, along with an optional SD map, as inputs. 
These inputs are initially combined to create BEV features. 
Subsequently, our method identifies lane centerlines as geometric points on the BEV features and determines their associations.
To enhance performance in intricate topological scenarios, we generate a mask for each centerline instance by leveraging the guidance of centerline points. 
This enables precise localization and refinement of the centerline instances.
The overall architecture is illustrated in Figure~\ref{fig:network}.

\subsection{SD Map Interaction}
\label{subsec:sdmap}

Applying BEV features extracted from online sensors have demonstrated potential for topology reasoning in computer vision tasks. Nevertheless, the presence of challenges such as occlusion and limited sensing range can result in inaccuracies. 
To address these issues, we present a novel SD map interaction module that effectively augments BEV features by incorporating beyond-visual-range data and road shape priors extracted from the SD map.
A piece of an SD map consists of multiple instances denoted as $s_1, s_2, \dots, s_{N_S}$, where $N_S$ represents the total number of SD map instances. 
Each instance $s_i$ is composed of a polyline represented by $\mathbf {sp}_i\in \mathbb R^{M_i\times 3}$, consisting of $M_i$ points, and a corresponding semantic type denoted as $st_i\in {1,2,\dots,N_M}$. 
Here, $N_M$ denotes the total number of semantic types available in the SD map.
We begin by converting the vectorized SD map instances into the BEV features. 
Each grid cell is filled with the corresponding semantic type embedding if it is occupied. 
Otherwise, a default semantic type embedding is used. 
This is represented as $\mathbf {E}_S\in \mathbb R^{HW\times C}$, where $HW$ is the grid dimensions and $C$ is the number of embedding dimensions.
The augmented BEV feature map, denoted as $\hat{\mathbf B}\in \mathbb R^{HW\times C}$, is obtained using the following equation:
\begin{equation}
    \hat{\mathbf B} = \text{TrDec}(\mathbf B, \mathbf E_S + \mathbf E_P),
\end{equation}
where $\mathbf B$ represents the BEV features obtained from online sensors, $\mathbf {E}_P\in \mathbb R^{HW\times C}$ is a positional encoding map for the SD grid, generated using sine and cosine functions of different frequencies \cite{transformer}. 
The function $\text{TrDec}$ refers to a transformer decoder that incorporates multi-layer deformable self- and cross-attentions among vanilla BEV feature map $\mathbf B$ and SD BEV feature map $\mathbf E_S+\mathbf E_P$. 
\subsection{Centerline Instance Prediction}
\label{subsec:detection}

\subsubsection{Centerline Detection.}
\label{subsubsec:det}

We begin by a BEV feature map, encoded with position embeddings, denoted as $\mathbf B\in \mathbb R^{HW\times C}$.
When using the SD map, $\mathbf B$ is replaced with $\hat{\mathbf B}$.
Here, $HW$ represents the size of the BEV space. 
We utilize a transformer decoder to extract centerline instances. 
Specifically, we incorporate a hybrid attention layer that combines masked cross-attention, deformable cross-attention, and self-attention. The masked cross-attention focuses on aggregating the features on the mask generated in Section \ref{subsec:optimization}. 
The deformable cross-attention aggregates the features of the learnable sampled points. 
Lastly, the self-attention facilitates interactions among centerline instance queries.
We denote a collection of learnable centerline instance queries as $\mathbf Q\in \mathbb R^{N_L\times C}$, where $N_L$ is the number of predicted centerline instances.
The masked cross-attention is defined by:
\begin{equation}
    \text{MaskedCrossAttn}(\mathbf Q, \mathbf B) = \text{softmax}(\mathbf M+\mathbf Q\mathbf B^{\text{T}})\mathbf B,
\end{equation}
where $\mathbf M\in \{0,-\infty\}^{N_L\times HW}$ represents the attention masks of instance queries $\mathbf Q$ over $\mathbf B$.
For more detailed information, please refer to the reference \cite{mask2former}.

Note that the driving scene encompasses two distinct types of centerlines: real and virtual. 
Real centerlines are situated in the middle of real lanes and possess clearly visible lane boundaries, while virtual centerlines mainly occur in intersections and serve as connections between distant lanes. 
To address this difference, we propose a real-virtual separation (RVS) strategy to independently process the real and virtual centerlines.
Firstly, we present two separate instance queries for real and virtual centerline learning. 
This allows us to focus on capturing the unique characteristics of each type. 
Secondly, we observe that the positions of virtual centerlines are dependent on real centerlines, while the positions of real centerlines have less reliance on virtual ones. 
To leverage this observation, we introduce a customized real-virtual separation-based self-attention module.
The real and virtual centerline queries are represented as $\mathbf Q^r\in \mathbb R^{N_R\times C}$ and $\mathbf Q^v\in \mathbb R^{N_V\times C}$, respectively. 
Here, $N_R$ and $N_V$ denote the number of real and virtual queries, respectively, with the constraint that $N_R+N_V=N_L$.
We implement the real-virtual separation-based self-attention by the following equation:
\begin{equation}
    \text{RVSelfAttn} = \text{softmax}\Biggl(\frac{\begin{bmatrix}\mathbf Q^r\mathbf Q^r{^{\text{T}}} & -\infty \\ \mathbf Q^v\mathbf Q^r{^{\text{T}}} & \mathbf Q^v\mathbf Q^v{^{\text{T}}}\end{bmatrix}}{\sqrt{C}}\Biggr)\begin{bmatrix}\mathbf Q^r\\ \mathbf Q^v\end{bmatrix}.
\end{equation}
After applying the hybrid attention and real-virtual separation of centerline decoder, we feed the centerline queries into a detection head.
The detection head consists of two multilayer perceptrons (MLP).
One MLP is responsible for generating two sets of centerline instances, $\mathbf L_R=\{\mathbf l_1,\mathbf l_2,\dots,\mathbf l_{N_R}\}$ and $\mathbf L_V=\{\mathbf l_{N_R+1},\dots,\mathbf l_{N_L}\}$.
The other MLP is used to generate the probabilities of centerlines, $\mathbf S\in \mathbb [0,1]^{N_L}$.
Each centerline instance $\mathbf l_i\in \mathbb R^{K\times 3}$ represents a polyline in the BEV space composed of $K$ fixed-order points.
During the training, the predicted real and virtual instances are separately handled through bipartite matching and loss construction, which are described in detail in Section \ref{subsec:loss}.

\subsubsection{Topological Association.}
\label{subsubsec:topoasso}

To achieve accurate topological association, we design a topology head based on centerline instance queries and positional embeddings. 
This head utilizes a classifier to predict a topological matrix $\mathbf A_{ll}\in [0,1]^{N_L\times N_L}$.
We first enhance the centerline instance queries with positional embeddings. 
The $i\text{-th}$ centerline instance query $\mathbf Q_i$ is enhanced using the following equation:
\begin{equation}
{\mathbf E}_i=\psi_1(\mathbf Q_i)+\psi_2(\mathbf l_i),
\end{equation}
where $\mathbf l_i\in \mathcal L$ represents the $i\text{-th}$ predicted centerline instance, $\psi_2:\mathbb R^{K\times 3}\rightarrow \mathbb R^{C}$ is an MLP layer that encodes positional information, and $\psi_1$ represents another MLP.
Next, the enhanced queries $\mathbf E\in \mathbb R^{N_L\times C}$ are repeated along a new axis, resulting in ${\mathbf \Lambda}_{ll}\in \mathbb R^{N_L\times N_L \times 2C}$, and then fed into a binary classifier to obtain the final topological matrix $\mathbf A_{ll}$.

\subsection{Points-Mask Optimization}
\label{subsec:optimization}

The centerline instance points obtained from Section \ref{subsec:detection} often lack sufficient accuracy, especially in cases with high curvature, as shown by those blue dots in Fig.~\ref{fig:teaser}. 
To address this issue, we propose a points-mask optimization module for the refinement of centerline instance points, consisting of two submodules: points-guided mask generation and points-mask fusion.
The former submodule is responsible for generating centerline instance mask, guided by the centerline points, while the latter submodule further refines the centerline instance points (represented as green dots in Fig.~\ref{fig:teaser}), based on the predictions from the mask.
It is important to note that virtual centerlines lack significant visual information, which makes the predicted mask for them less accurate for refining the centerline points.
Hence, for virtual centerlines, we solely generate a mask for supervision, without refining their points based on the mask. 
However, for real centerlines, we obtain refined points denoted as $\mathbf L_{R}^{\prime}=\{\mathbf l_1^{\prime},\dots, \mathbf l_{N_R}^{\prime}\}$.

\subsubsection{Points-Guided Mask Generation.}
\label{subsubsec:maskgen}

Mask2Former\cite{mask2former} proposed a DETR-like instance segmentation architecture that relies on instance mask queries.
However, the mask queries used in Mask2Former are learnable embeddings without positional priors.
To address this limitation and enhance the quality of the generated mask, we propose a points-guided mask generation approach that leverages the detected centerline points to guide the generation of mask queries.
Specifically, we employ a positional encoding layer and a query encoding layer to encode the centerline points $\mathbf l_i\in \mathbb R^{K\times 3}$ and the query $\mathbf Q_i$, respectively. 
We then combine them by summing the positional encodings and query encodings to obtain the mask query $\mathbf Q^{\prime}_i$.
In the positional encoding layer, each point in $\mathbf l_i$ is encoded using a shared MLP, and the resulting encodings are concatenated and processed by another MLP to obtain the positional embedding $\mathbf f_i\in \mathbb R^{C}$. 
The query encoding layer is also implmented by an MLP.
Finally, we input the mask query $\mathbf Q^{\prime}_i$ and the BEV features $\mathbf B$ into a dot product-based mask head to generate the mask $\mathbf M_i\in \mathbb R^{HW}$ for the $i$-th centerline:
\begin{equation}
    \mathbf M_i = \mathbf B \cdot \mathbf Q^{\prime}_i,
\end{equation}

\subsubsection{Points-Mask Fusion.}

After obtaining the mask for each centerline instance in the BEV space, we aim to refine real centerlines by incorporating information from the corresponding mask. 
However, directly integrating the mask and centerline points is not trivial due to their inherently distinct representations.
To address this challenge, we propose a two-stage process for blending the mask and points. 
The first stage, referred to as mask points sampling, involves sampling a set of mask points from the generated mask. 
In the second stage, known as points fusion, we combine the detected centerline points obtained from Section \ref{subsec:detection} with the sampled mask points to obtain refined centerline points. 

In the mask points sampling stage, we aim to select a fixed number of points from each mask instance $\mathbf M_i$. 
To achieve this, we regress one point for each column, resulting in a set of $W$ ordered points. 
The position of each sampling mask point in the $j$-th column of the $i$-th centerline instance, denoted as $\mathbf {C}_{i,j}$, can be calculated using the equation:
\begin{equation}
    \mathbf C_{i,j} = [0,1,\dots, H-1]^{\text{T}} \cdot \text{softmax}(\mathbf M_i(:,j)),
\end{equation}
where $\mathbf M_i(:,j)$ refers to the column $j$ of the mask $\mathbf M_i$.
Additionally, since the length of the centerline instance is finite and the direction of mask points is unknown (whether it is left-to-right or right-to-left), we also predict the existence probabilities of the mask points $\mathbf P_i$ (in the range of $[0,1]$) and the left-to-right direction probability $D_i$ (also in the range of $[0,1]$). 
The existence probabilities of the mask points and left-to-right direction probability can be calculated by:
\begin{equation}
    \begin{split}
        \mathbf P_i&=\text{sigmoid}(\phi_1(\mathbf M_i)),\\
                D_i&=\text{sigmoid}(\phi_2(\mathbf Q^{\prime}_i)),
    \end{split}
\end{equation}
where $\phi_1:\mathbb R^{HW}\rightarrow \mathbb R^{W}$ is an MLP that transforms the mask to the dimension of $W$, and $\phi_2\in \mathbb R^C\rightarrow \mathbb R$ is another MLP.
It is worth noting that the above mask point sampling approach cannot handle centerlines that are completely vertical, as they contain multiple points in a single column. 
To address this issue, we regress one point per row to generate another set of points for vertical lines. 
During training, bipartite graph matching and loss construction are performed separately for the two sets of point sets.
During the testing, we select the set of points with a higher number of valid mask points for evaluation, where mask points are considered valid if their existence probabilities are greater than 0.5.

In the points fusion stage, we begin by choosing valid mask points and filtering out outliers among them. 
An outlier is defined as a mask point whose minimum distance to its neighboring mask points exceeds 1.5m. 
Next, we perform resampling on the valid mask points, generating a new set containing $K$ points. 
We then take the average of these resampled points with the detected centerline points $\mathbf l_i$ obtained from Section~\ref{subsec:detection}. 
This averaging process yields the final refined points $\mathbf l^{\prime}_i$ in $\mathbb R^{K\times 3}$.
 
\subsection{Loss Function}
\label{subsec:loss}

We categorize the ground truth centerlines into two categories: real and virtual. 
To construct losses, we perform bipartite matching separately for the virtual centerlines predicted in Section~\ref{subsec:detection} and the real centerlines obtained from Section~\ref{subsec:optimization} with each category of ground truth. 
The final form of our loss function is presented as follows:
\begin{equation}
    \mathcal L = \mathcal L_{\text{top}}(\mathbf A_{ll}) + \mathcal L_{\text{cls}}(\mathbf S) + \mathcal L_{\text{det}}(\mathbf L_V,\mathbf L_R) + \mathcal L_{\text{mask}}(\mathbf M) + \mathcal L_{\text{mp}}(\mathbf C,\mathbf P,D),
    \label{equ:loss}
\end{equation}
where we only descirbe the predictions in the loss function without explicitly mentioning the ground truth for brevity.
The first two losses, denoted as $\mathcal L_{top}(\mathbf A_{ll})$ and $\mathcal L_{\text{cls}}(\mathbf S)$, are implemented using focal loss. 
These losses are utilized to supervise the topological associations and centerline probabilities, respectively.
The detection term, denoted as $\mathcal L_{\text{det}}(\mathbf L_V,\mathbf L_R)$, is computed using the L1 loss, which is employed to supervise the geometric shape of centerlines.
To supervise the generation of the mask $\mathbf M$ for both real and virtual centerlines, we use the loss function $\mathcal L_{\text{mask}}(\mathbf M)$, which combines the binary cross-entropy loss and the dice loss\cite{diceloss}. 
Furthermore, we employ the L1 loss, the binary cross-entropy loss, and the focal loss to supervise the column-based sampling mask points $\mathbf C$, the existence probabilities $\mathbf P$, and the points direction $D$ in the function $\mathcal L_{mp}(\cdot)$, respectively.
In addition, we also add supervision for the row-based sampling mask points, following the same approach as the column-based ones. However, for brevity, this supervision is not explicitly stated in the equation.

\section{Experiments}
\label{section:exp}

\subsection{Dataset and Metrics}
\subsubsection{Dataset.} 
We conducted all experiments on the OpenLane-V2 dataset\cite{OpenLaneV2}.
The dataset consists of 1000 scenes captured in various autonomous driving scenarios, each including multiview images and several annotations, such as centerlines, traffic elements, and topology relationships, sampled at 2Hz.
Centerlines are defined by 201 ordered points in 3D space, spanning the range of $[-50m, +50m]$ along the x-axis and $[-25m, +25m]$ along the y-axis. 
The traffic elements, categorized into thirteen classes, are annotated using 2D bounding boxes in the front-view images.
Moreover, the dataset provides adjacency matrices to represent two types of topology relationships: centerline-centerline and centerline-traffic element relationships.

\subsubsection{Evaluation Metrics.}
The evaluation encompasses detection and topology metrics. 
The detection scores for centerlines ($\text{DET}_l$) and traffic elements ($\text{DET}_t$) are computed as mean average precision (mAP), utilizing Fr\'echet distance and Intersection over Union (IoU), respectively.
The topological scores for centerline-centerline relationships ($\text{TOP}_{ll}$) and centerline-traffic element relationships ($\text{TOP}_{lt}$) are conducted using mAP metrics grounded in graph theory principles.

\subsection{Implementation Details}
\subsubsection{Model Details.}
Following the settings in TopoNet\cite{TopoNet}, we utilize the ResNet-50\cite{resnet50} as image backbone pretrained on ImageNet\cite{imagenet} to extract multi-scale image features. 
The multi-scale features $\{\mathbf{I}_{3}, \mathbf{I}_{4}, \mathbf{I}_{5}\}$ from the last three stages of ResNet-50 are employed for constructing BEV features.
For PV-to-BEV transformation, we follow the default settings of BEVFormer\cite{bevformer} with $3$ encoder layers to integrate the multi-scale features.
The centerline detector consists of $4$ decoder layers, each with $8$ attention heads and $C=256$ channels.
The dimension of feedforward network is set to $512$.
We define the number of real centerline queries and virtual centerline queries as $N_{R}=150$ and $N_{V}=150$, respectively. 
We opt for an 11 fixed-order points representation for $K$ in Section \ref{subsec:detection}.
For fair comparison, we adopt the same network configuration as TopoNet for extracting traffic elements and reasoning lane centerline and traffic element topology.

\subsubsection{Training.}
We resize the input image to $1024 \times 800$ and apply the same image data augmentation as in \cite{TopoNet}. 
The loss coefficients are configured as follows: $\lambda_\text{top} = 5$, $\lambda_\text{cls} = 1.5$, $\lambda_\text{det} = 0.025$, $\lambda_\text{mask} = 1$, and $\lambda_\text{mp} = 7$.
We utilize the AdamW optimizer\cite{adamw} with an initial learning rate of $2e^{-4}$. 
The learning rates for the backbone are set to one-tenth of those for other modules. 
Our model is trained for 24 epochs using cosine annealing with a weight decay of 0.01.
To prevent gradient explosion, we apply gradient clipping during backpropagation, limiting the maximum norm to 35.

\begin{table}[!b]
    \caption{Comparison with SOTA methods on the OpenLane-V2 \textit{subset\_A} dataset. $*$ indicates methods using SD map. The best results are bolded.
    }
    \label{tab:openlanev2}
    \centering
    \begin{tabular}{c|cc|ccccc}
        \toprule
        {Method} & {Backbone} & {epoch} &
        {$\text{DET}_{l}\uparrow$} & {$\text{DET}_t\uparrow$} & {$\text{TOP}_{ll}\uparrow$} & {$\text{TOP}_{lt}\uparrow$} & {$\text{OLS}\uparrow$} \\
        \midrule
        STSU \cite{STSU}
        & {ResNet-50} & 24
        & 12.7 & 43.0
        & 0.5 & 15.1
        & 25.4 \\
        VectormapNet \cite{VectorMapNet}
        & {ResNet-50} & 24
        & 11.1 & 41.7
        & 0.4 & 5.9
        & 20.8 \\
        MapTR \cite{MapTR}
        & {ResNet-50} & 24
        & 17.7 & 43.5
        & 1.1 & 10.4
        & 26.0 \\
        TopoNet \cite{TopoNet}
        & {ResNet-50} & 24
        & 28.5 & 48.1
        & 4.1 & 20.8
        & 35.6 \\ 
        TopoMLP \cite{TopoMLP}
        & {ResNet-50} & 24
        & 28.3 & \textbf{50.0}
        & 7.2  & 22.8 
        & 38.2 \\
        RoadPainter 
        & {ResNet-50} & 24
        & \textbf{30.7} & 47.7
        & \textbf{7.9} & \textbf{24.3}
        & \textbf{38.9}  \\
        \midrule
        $\text{SMERF}^{*}$ \cite{SMERF}
        & {ResNet-50} & 24
        & 33.4 & \textbf{48.6}
        & 7.5 & 23.4
        & 39.4 \\ 
        $\text{RoadPainter}^{*}$
        & {ResNet-50} & 24
        & \textbf{36.9} & 47.1
        & \textbf{12.7} & \textbf{25.8}
        & \textbf{42.6} \\
        \bottomrule
    \end{tabular}
\end{table}
\begin{table}[!b]
    \caption{Comparison with SOTA methods on the OpenLane-V2 \textit{subset\_B} dataset.
    }
    \label{tab:openlanev2_b_main}
    \centering
    \begin{tabular}{c|cc|ccccc}
        \toprule
        {Method} & {Backbone} & {epoch} &
        {$\text{DET}_{l}\uparrow$} & {$\text{DET}_t\uparrow$} & {$\text{TOP}_{ll}\uparrow$} & {$\text{TOP}_{lt}\uparrow$} & {$\text{OLS}\uparrow$} \\
        \midrule
        STSU\cite{STSU}
        & {ResNet-50} & 24
        & 8.2 & 43.9
        & 0.0 & 9.4
        & 21.2 \\
        VectormapNet\cite{VectorMapNet}
        & {ResNet-50} & 24
        & 3.5 & 49.1
        & 0.0 & 1.4
        & 16.3 \\
        MapTR\cite{MapTR}
        & {ResNet-50} & 24
        & 15.2 & 54.0
        & 0.5 & 6.1
        & 25.2 \\
        TopoNet\cite{TopoNet}
        & {ResNet-50} & 24
        & 24.3 & 55.0
        & 2.5 & 14.2
        & 33.2 \\ 
        TopoMLP\cite{TopoMLP}
        & {ResNet-50} & 24
        & 26.6 & \textbf{58.3}
        & 7.6  & \textbf{17.8}
        & \textbf{38.7} \\
        \midrule
        RoadPainter 
        & {ResNet-50} & 24
        & \textbf{28.7} & 54.8
        & \textbf{8.5} & 17.2
        & 38.5  \\
        \bottomrule
    \end{tabular}
\end{table}

\subsection{Comparison with State-of-the-Arts}
We conduct a comparative evaluation of RoadPainter against current state-of-the-art (SOTA) methods on the OpenLane-V2 dataset\cite{OpenLaneV2}.
The results are reported in Table \ref{tab:openlanev2} and Table \ref{tab:openlanev2_b_main}, where we ensure a fair comparison by employing the same backbone for all methods.
Our proposed method, RoadPainter, achieves an OLS score of 38.9, surpassing the performance of other SOTA methods on \textit{subset\_A} dataset.
In comparison to alternative approaches, RoadPainter demonstrates superior results in terms of $\text{DET}_l$ (+2.2) and $\text{TOP}_{ll}$ (+0.7) metrics, highlighting the effectiveness of our model design.
We also conducted experiments on the OpenLane-V2 \textit{subset\_B} dataset.
Our approach outperforms others in centerline detection $\text{DET}_{l}$ and topology $\text{TOP}_{ll}$. 
We employ a 3-layer BEVFormer architecture, similar to TopoNet, while TopoMLP utilizes a 6-layer PETR\cite{petr} architecture. 
The PETR architecture is particularly suitable for traffic sign detection and achieves superior results on metrics such as $\text{DET}_{t}$ and $\text{TOP}_{lt}$.
Moving forward, our main focus will be on enhancing traffic sign detection and topology association.
Furthermore, by incorporating the SD map, our enhanced model, RoadPainter$^*$, exhibits notable improvements in $\text{DET}_l$ (+6.2), $\text{TOP}_{ll}$ (+4.8) and OLS (+3.7) metrics. 
RoadPainter$^*$ notably outperforms SMERF$^{*}$ in both centerline detection and topology reasoning performance by a large margin.
The OpenLane-V2 \textit{subset\_B} dataset does not include an SD map, so there are no available results for $\text{RoadPainter}^{*}$. 
The FPS of our model (FP32) on the RTX3090 is 6.5.

\subsection{Ablation Studies}
We conduct ablation studies on the point-guided mask generation (PGM), point-mask fusion (PMF), and SD map interaction (SD) modules using the \textit{subset\_A} dataset. 
Our analysis focuses on quantifying the impact of each module, with the detailed improvements outlined in Table \ref{tab:ablation}.
In addition, we extend our assessment to include the instance segmentation performance of centerline. 
This evaluation is based on the AP (average precision) metric, represented as $\text{AP}_{l}$.

\textbf{Baseline.}
We construct the baseline model by retaining the vanilla centerline decoder without hybrid attention and real-virtual separation. 
In this setup, we utilize self-attention and deformable cross-attention mechanisms while omitting masked cross-attention.
For supervision, we only incorporate the first three losses as outlined in Equation~\ref{equ:loss}. 
Our baseline model achieves an OLS score of 37.2. 
Notably, the $\text{TOP}_{ll}$ metric of the baseline model (7.7) surpasses that of TopoNet (4.1), underscoring the efficacy of our topological association design.

\textbf{Point-Guided Mask Generation.}
Building upon the baseline model, we incorporate the PGM module to infer centerline masks guided by the regressed points.
When comparing the PGM model to the baseline, we observe a notable improvement of $1.2$ in $\text{DET}_{l}$ and $0.2$ in $\text{TOP}_{ll}$.
This significant enhancement can be attributed to our innovative design of point-guided mask generation.

\textbf{Point-Mask Fusion.}
We introduce a point-mask fusion module to enhance the accuracy of real centerlines by refining the points based on the generated masks.
We observe improvements in the $\text{DET}_{l}$, OLS, and $\text{AP}_{l}$ metrics, with increases of $2.6$, $1.3$, and $0.6$, respectively. 
These results highlight the efficacy of the PMF module in refining real centerlines, thereby benefiting centerline detection, topology prediction, and mask generation.

\textbf{SD Map Interaction.}
We employ the SD map interaction module to enhance the BEV features, thereby improving the accuracy of centerline detection and topological reasoning.
By incorporating information from the SD map, our final model exhibits significant performance gains compared to previous models that do not utilize the SD map. 
Specifically, there is a substantial improvement in $\text{DET}_{l}$ (+6.2) and $\text{TOP}_{ll}$ (+4.8).

\begin{table}[!b]
    \caption{The ablation studies investigate the effectiveness of the point-guided mask generation (PGM), point-mask fusion (PMF), and SD map interaction (SD) modules.
    }
    \label{tab:ablation}
    \centering
    \begin{tabular}{ccc|cccccc}
        \toprule
        {PGM} & 
        {PMF} & {SD} &
        {$\text{DET}_{l}\uparrow$} & {$\text{TOP}_{ll}\uparrow$} & {$\text{DET}_t\uparrow$} & {$\text{TOP}_{lt}\uparrow$} & {$\text{OLS}\uparrow$} & {$\text{AP}_{l}\uparrow$} \\
        \midrule
        & & &
        26.9 & 7.7  & 46.8 & 22.4 & 37.2 & \text{-} \\
        $\checkmark$ &  &
        & 28.1 & 7.9 & 46.6 & 22.6 & 37.6 & 13.5 \\
        $\checkmark$ & $\checkmark$ &
        & 30.7 & 7.9 & \textbf{47.7} & 24.3 & 38.9 & 14.1 \\
        $\checkmark$ & $\checkmark$ & $\checkmark$
        & \textbf{36.9} & \textbf{12.7} & 47.1 & \textbf{25.8} & \textbf{42.6} & \textbf{15.4} \\ 
        \bottomrule
    \end{tabular}
\end{table}
\begin{table}[!b]
    \caption{Ablations on attention mechanisms and segmentation masks.
    }
    \label{tab:ablation23_mix}
    \centering
    \begin{tabular}{ccc}
        \toprule
        {Method} & 
        {$\text{DET}_{l}\uparrow$} & {$\text{TOP}_{ll}\uparrow$} \\
        \midrule
        RoadPainter
        & \textbf{30.7} & \textbf{7.9}  \\
        \midrule
        w/o hybrid attention
        & 29.6 & 7.2 \\
        w/o real-virtual self-attn
        & 29.6 & 7.5 \\
        \midrule
        Only detection
        & 26.9 & 7.7  \\
        Segmentation as auxiliary supervision
        & 28.1 & 7.9  \\
        Segmentation as centerline sampling
        & 29.4 & 7.7 \\
        \bottomrule
    \end{tabular}
\end{table}

To better understand the attention mechanism in our method, we exclude masked attention in the transformer decoder (w/o hybrid attention) and omit real-virtual separation-based self-attention strategy (w/o real-virtual self-attn).
As depicted in Table~\ref{tab:ablation23_mix}, 
the elimination of hybrid attention results in a slight performance decrease across all metrics.
Furthermore, the absence of the real-virtual self-attention strategy notably affects the $\text{TOP}_{ll}$ metric (-0.4), indicating the contribution of the real-virtual separation-based self-attention approach.

To further verify the segmentation mask design in our method, we conduct three ablatin experiments: only detection, segmentation as auxiliary supervision, and segmentation as centerline sampling, as shown in Table \ref{tab:ablation23_mix}. 
The findings indicate that segmentation, when used independently for auxiliary supervision and mask point sampling, can enhance the performance of centerline detection.
Our method amalgamates these two benefits, thereby achieving superior results.

\subsection{Qualitative Results Analysis}
\begin{figure}[!h]
	\centering
	\includegraphics[width=\linewidth]{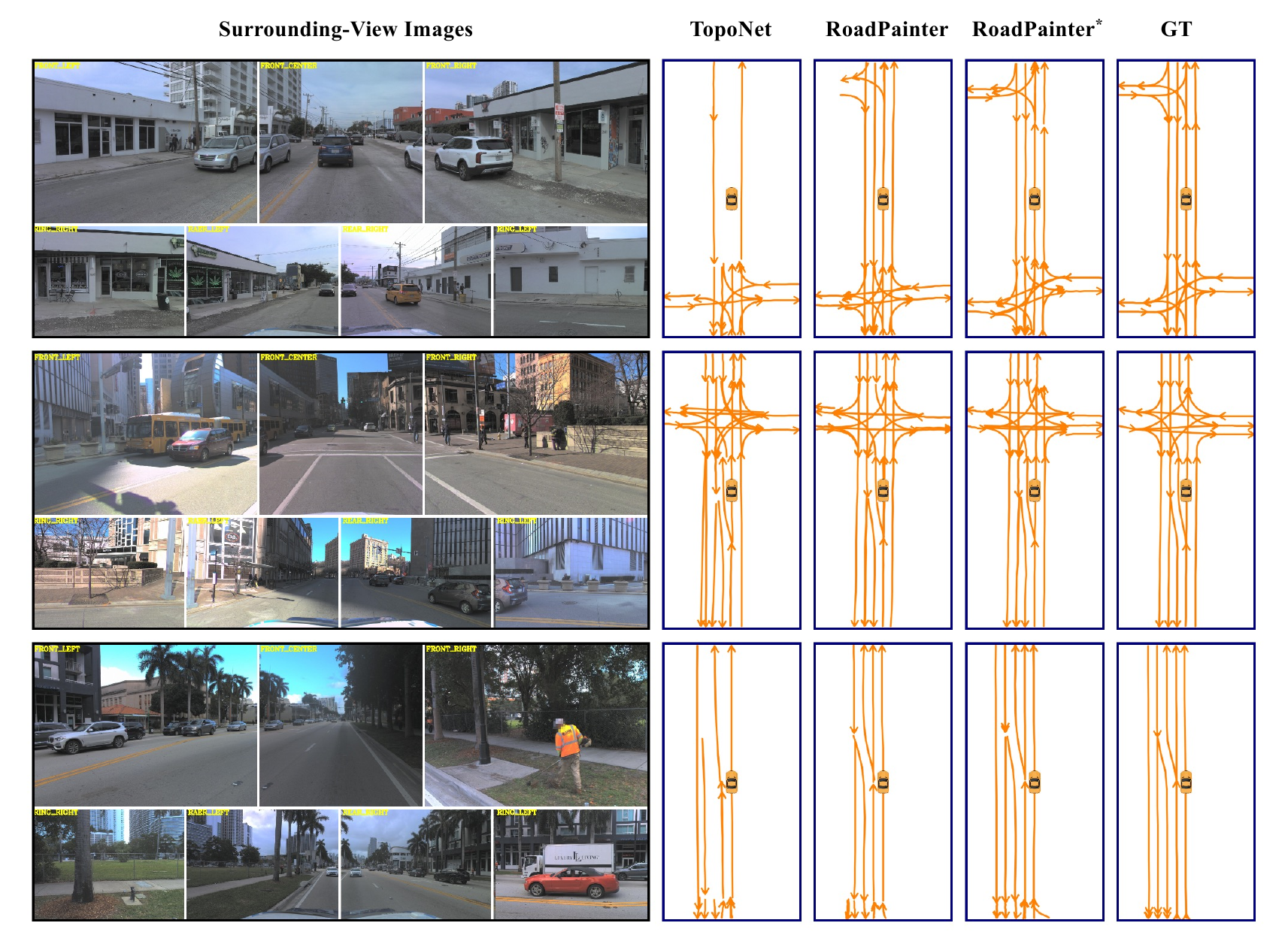}
	\caption{
		Qualitative evaluation of RoadPainter and comparison methods.
		Given multi-view images, RoadPainter achieves superior centerline detection performance compared to TopoNet in terms of completeness and accuracy.
		With the design of SD map interaction module, RoadPainter$^{*}$ precisely estimates the lane count at intersections.
	}
	\label{fig:vis_compare}

\end{figure}

\begin{figure}[tb]
	\centering
	\includegraphics[width=\linewidth]{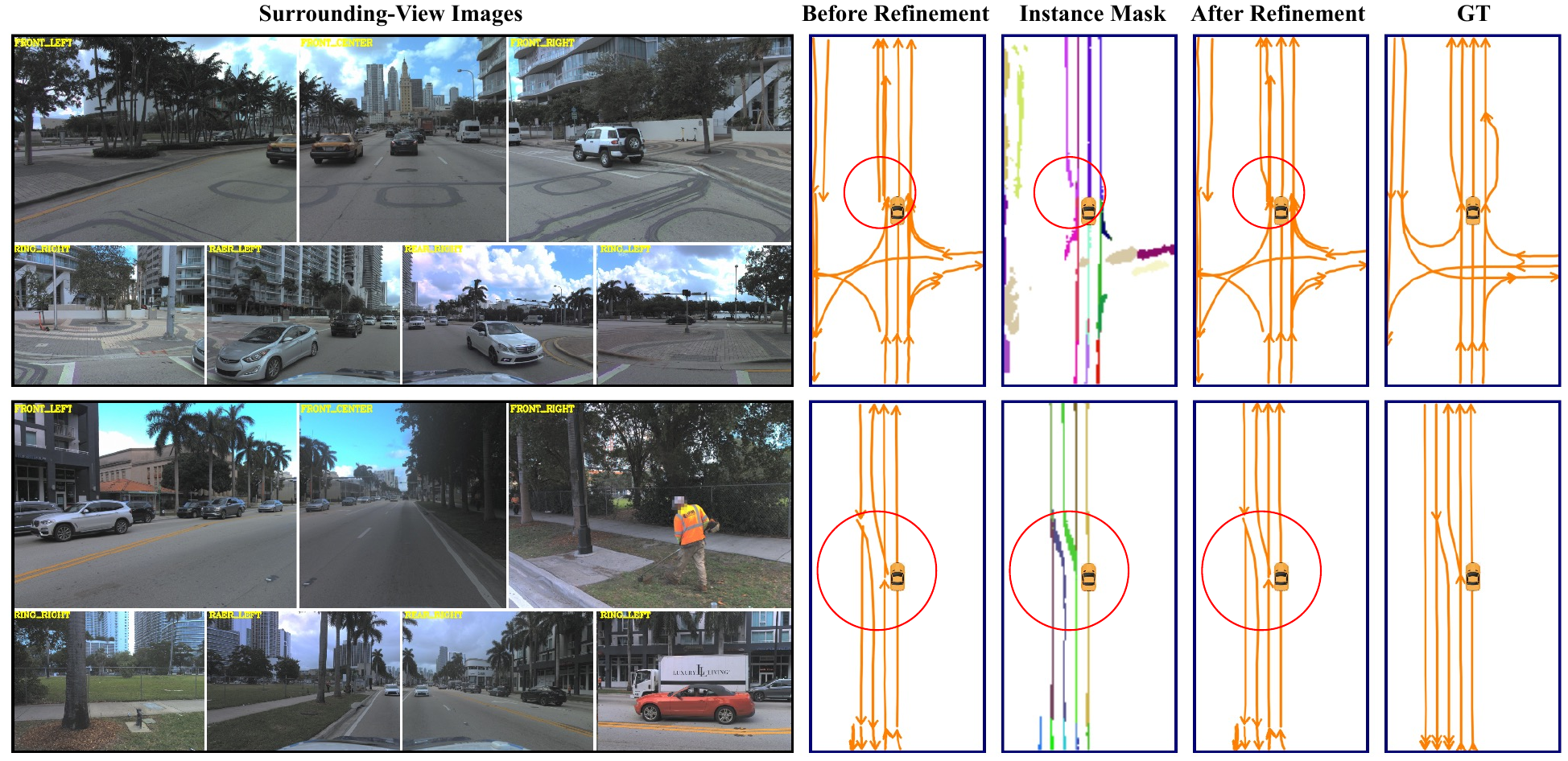}
	\caption{
		Qualitative analysis of instance mask of RoadPainter. 
		The directly regressed centerlines from the transformer decoder, along with the instance masks and the refined centerlines resulting from the points-mask optimization module, are presented.
	}
	\label{fig:vis_refine}
\end{figure}

\begin{figure}[tb]
	\includegraphics[width=\linewidth]{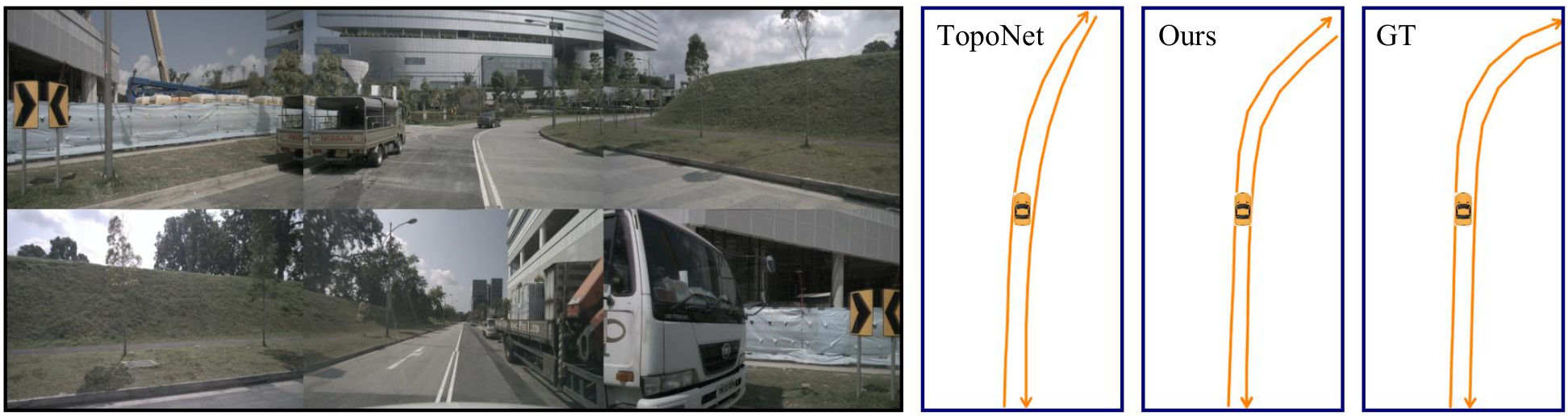}
	\caption{
		Our method outperforms TopoNet in detecting high-curvature lanes due to the utilization of segmentation masks, which provide more precise geometric details.
	}
	\label{fig:rebuttal_vis}
\end{figure}

We present a comparison of centerline visualization results among TopoNet, RoadPainter, RoadPainter$^{*}$ (with SD map), and ground truth (GT) in Fig.~\ref{fig:vis_compare}.
The findings illustrate that RoadPainter exhibits superior lane detection and topological reasoning capabilities, particularly in complex scenarios such as lane transitions and intersections.
RoadPainter$^{*}$ demonstrates the ability to accurately predict centerlines and topological relationships beyond the visual range, facilitated by the SD map interaction module.
In Fig.~\ref{fig:vis_refine}, we further present visualization results of centerlines from Section~\ref{subsec:detection}, along with instance masks and refined centerlines from Section~\ref{subsec:optimization}.
These results clearly demonstrate that, under the guidance of the instance mask, the refined centerlines exhibit superior performance in terms of accuracy, particularly at the junctions where details are crucial.
In Fig.~\ref{fig:rebuttal_vis}, our method exhibits superior performance compared to TopoNet in accurately extracting high-curvature lanes, which can be attributed to the utilization of segmentation masks. 
These findings indicate that our method brings notable improvements in lane detection, particularly in the accurate extraction of high-curvature lanes.

\section{Conclusion}
\label{section:conclusion}

We introduced RoadPainter, a cutting-edge end-to-end network tailored for centerline detection and topological reasoning, which enhances performance through points-guided mask generation and mask-based centerline refinement. 
We designed a transformer decoder that incorporates hybrid attention and real-virtual separation strategies to regress lane centerline points and reason topological relationships. 
Furthermore, we developed a centerline points-guided mask generation module that generates centerline masks guided by centerline points to refine the centerline detection process. 
We further refined the centerline points by integrating centerline masks through a points-mask fusion module, eliminating the necessity of post-processing. 
The experimental evaluation on the OpenLane-V2 dataset demonstrated that RoadPainter achieves superior performance, validating the effectiveness of our module designs.

\bibliographystyle{splncs04}
\bibliography{main}

\begin{thebibliography}{10}
\providecommand{\url}[1]{\texttt{#1}}
\providecommand{\urlprefix}{URL }
\providecommand{\doi}[1]{https://doi.org/#1}

\bibitem{STSU}
Can, Y.B., Liniger, A., Paudel, D.P., Van~Gool, L.: Structured bird’s-eye-view traffic scene understanding from onboard images. In: Proceedings of the IEEE/CVF International Conference on Computer Vision. pp. 15661--15670 (2021)

\bibitem{TPLR}
Can, Y.B., Liniger, A., Paudel, D.P., Van~Gool, L.: Topology preserving local road network estimation from single onboard camera image. In: Proceedings of the IEEE/CVF Conference on Computer Vision and Pattern Recognition. pp. 17263--17272 (2022)

\bibitem{OLC}
Can, Y.B., Liniger, A., Paudel, D.P., Van~Gool, L.: Improving online lane graph extraction by object-lane clustering. In: Proceedings of the IEEE/CVF International Conference on Computer Vision. pp. 8591--8601 (2023)

\bibitem{PersFormer}
Chen, L., Sima, C., Li, Y., Zheng, Z., Xu, J., Geng, X., Li, H., He, C., Shi, J., Qiao, Y., Yan, J.: Persformer: 3d lane detection via perspective transformer and the openlane benchmark. In: European Conference on Computer Vision (ECCV) (2022)

\bibitem{mask2former}
Cheng, B., Misra, I., Schwing, A.G., Kirillov, A., Girdhar, R.: Masked-attention mask transformer for universal image segmentation (2022)

\bibitem{Pivotnet}
Ding, W., Qiao, L., Qiu, X., Zhang, C.: Pivotnet: Vectorized pivot learning for end-to-end hd map construction. In: Proceedings of the IEEE/CVF International Conference on Computer Vision. pp. 3672--3682 (2023)

\bibitem{3D-LaneNet}
Garnett, N., Cohen, R., Pe'er, T., Lahav, R., Levi, D.: 3d-lanenet: End-to-end 3d multiple lane detection. In: Proceedings of the IEEE/CVF International Conference on Computer Vision (ICCV) (October 2019)

\bibitem{resnet50}
He, K., Zhang, X., Ren, S., Sun, J.: Deep residual learning for image recognition. In: Proceedings of the IEEE conference on computer vision and pattern recognition. pp. 770--778 (2016)

\bibitem{st-p3}
Hu, S., Chen, L., Wu, P., Li, H., Yan, J., Tao, D.: St-p3: End-to-end vision-based autonomous driving via spatial-temporal feature learning. In: European Conference on Computer Vision. pp. 533--549. Springer (2022)

\bibitem{uniad}
Hu, Y., Yang, J., Chen, L., Li, K., Sima, C., Zhu, X., Chai, S., Du, S., Lin, T., Wang, W., et~al.: Planning-oriented autonomous driving. In: Proceedings of the IEEE/CVF Conference on Computer Vision and Pattern Recognition. pp. 17853--17862 (2023)

\bibitem{imagenet}
Krizhevsky, A., Sutskever, I., Hinton, G.E.: Imagenet classification with deep convolutional neural networks. Advances in neural information processing systems  \textbf{25} (2012)

\bibitem{HDMapNet}
Li, Q., Wang, Y., Wang, Y., Zhao, H.: Hdmapnet: An online hd map construction and evaluation framework. In: 2022 International Conference on Robotics and Automation (ICRA). pp. 4628--4634. IEEE (2022)

\bibitem{TopoNet}
Li, T., Chen, L., Wang, H., Li, Y., Yang, J., Geng, X., Jiang, S., Wang, Y., Xu, H., Xu, C., Yan, J., Luo, P., Li, H.: Graph-based topology reasoning for driving scenes. arXiv preprint arXiv:2304.05277  (2023)

\bibitem{bevformer}
Li, Z., Wang, W., Li, H., Xie, E., Sima, C., Lu, T., Qiao, Y., Dai, J.: Bevformer: Learning bird’s-eye-view representation from multi-camera images via spatiotemporal transformers. In: European conference on computer vision. pp. 1--18. Springer (2022)

\bibitem{LaneGAP}
Liao, B., Chen, S., Jiang, B., Cheng, T., Zhang, Q., Liu, W., Huang, C., Wang, X.: Lane graph as path: Continuity-preserving path-wise modeling for online lane graph construction. arXiv preprint arXiv:2303.08815  (2023)

\bibitem{MapTR}
Liao, B., Chen, S., Wang, X., Cheng, T., Zhang, Q., Liu, W., Huang, C.: Maptr: Structured modeling and learning for online vectorized hd map construction. In: International Conference on Learning Representations (2023)

\bibitem{MapTRv2}
Liao, B., Chen, S., Zhang, Y., Jiang, B., Zhang, Q., Liu, W., Huang, C., Wang, X.: Maptrv2: An end-to-end framework for online vectorized hd map construction. arXiv preprint arXiv:2308.05736  (2023)

\bibitem{CondLaneNet}
Liu, L., Chen, X., Zhu, S., Tan, P.: Condlanenet: a top-to-down lane detection framework based on conditional convolution. In: Proceedings of the IEEE/CVF International Conference on Computer Vision. pp. 3773--3782 (2021)

\bibitem{VectorMapNet}
Liu, Y., Yuan, T., Wang, Y., Wang, Y., Zhao, H.: Vectormapnet: End-to-end vectorized hd map learning. In: International Conference on Machine Learning. pp. 22352--22369. PMLR (2023)

\bibitem{petr}
Liu, Y., Wang, T., Zhang, X., Sun, J.: Petr: Position embedding transformation for multi-view 3d object detection. arXiv preprint arXiv:2203.05625  (2022)

\bibitem{adamw}
Loshchilov, I., Hutter, F.: Decoupled weight decay regularization. arXiv preprint arXiv:1711.05101  (2017)

\bibitem{SMERF}
Luo, K.Z., Weng, X., Wang, Y., Wu, S., Li, J., Weinberger, K.Q., Wang, Y., Pavone, M.: Augmenting lane perception and topology understanding with standard definition navigation maps. arXiv preprint arXiv:2311.04079  (2023)

\bibitem{diceloss}
Milletari, F., Navab, N., Ahmadi, S.A.: V-net: Fully convolutional neural networks for volumetric medical image segmentation. In: 2016 fourth international conference on 3D vision (3DV). pp. 565--571. Ieee (2016)

\bibitem{wayformer}
Nayakanti, N., Al-Rfou, R., Zhou, A., Goel, K., Refaat, K.S., Sapp, B.: Wayformer: Motion forecasting via simple \& efficient attention networks. In: 2023 IEEE International Conference on Robotics and Automation (ICRA). pp. 2980--2987. IEEE (2023)

\bibitem{BeMapNet}
Qiao, L., Ding, W., Qiu, X., Zhang, C.: End-to-end vectorized hd-map construction with piecewise bezier curve. In: Proceedings of the IEEE/CVF Conference on Computer Vision and Pattern Recognition. pp. 13218--13228 (2023)

\bibitem{UFAST}
Qin, Z., Wang, H., Li, X.: Ultra fast structure-aware deep lane detection. In: Computer Vision--ECCV 2020: 16th European Conference, Glasgow, UK, August 23--28, 2020, Proceedings, Part XXIV 16. pp. 276--291. Springer (2020)

\bibitem{p3}
Sadat, A., Casas, S., Ren, M., Wu, X., Dhawan, P., Urtasun, R.: Perceive, predict, and plan: Safe motion planning through interpretable semantic representations. In: Computer Vision--ECCV 2020: 16th European Conference, Glasgow, UK, August 23--28, 2020, Proceedings, Part XXIII 16. pp. 414--430. Springer (2020)

\bibitem{mtr++}
Shi, S., Jiang, L., Dai, D., Schiele, B.: Mtr++: Multi-agent motion prediction with symmetric scene modeling and guided intention querying. IEEE Transactions on Pattern Analysis and Machine Intelligence  (2024)

\bibitem{InstaGram}
Shin, J., Rameau, F., Jeong, H., Kum, D.: Instagram: Instance-level graph modeling for vectorized hd map learning. arXiv preprint arXiv:2301.04470  (2023)

\bibitem{multipath++}
Varadarajan, B., Hefny, A., Srivastava, A., Refaat, K.S., Nayakanti, N., Cornman, A., Chen, K., Douillard, B., Lam, C.P., Anguelov, D., et~al.: Multipath++: Efficient information fusion and trajectory aggregation for behavior prediction. In: 2022 International Conference on Robotics and Automation (ICRA). pp. 7814--7821. IEEE (2022)

\bibitem{transformer}
Vaswani, A., Shazeer, N., Parmar, N., Uszkoreit, J., Jones, L., Gomez, A.N., Kaiser, {\L}., Polosukhin, I.: Attention is all you need. Advances in neural information processing systems  \textbf{30} (2017)

\bibitem{OpenLaneV2}
Wang, H., Li, T., Li, Y., Chen, L., Sima, C., Liu, Z., Wang, B., Jia, P., Wang, Y., Jiang, S., Wen, F., Xu, H., Luo, P., Yan, J., Zhang, W., Li, H.: Openlane-v2: A topology reasoning benchmark for unified 3d hd mapping. In: NeurIPS (2023)

\bibitem{GANet}
Wang, J., Ma, Y., Huang, S., Hui, T., Wang, F., Qian, C., Zhang, T.: A keypoint-based global association network for lane detection. In: Proceedings of the IEEE/CVF Conference on Computer Vision and Pattern Recognition (CVPR). pp. 1392--1401 (June 2022)

\bibitem{BEV-LaneDet}
Wang, R., Qin, J., Li, K., Li, Y., Cao, D., Xu, J.: Bev-lanedet: An efficient 3d lane detection based on virtual camera via key-points. In: Proceedings of the IEEE/CVF Conference on Computer Vision and Pattern Recognition. pp. 1002--1011 (2023)

\bibitem{LaneNet}
Wang, Z., Ren, W., Qiu, Q.: Lanenet: Real-time lane detection networks for autonomous driving. arXiv preprint arXiv:1807.01726  (2018)

\bibitem{TopoMLP}
Wu, D., Chang, J., Jia, F., Liu, Y., Wang, T., Shen, J.: Topomlp: An simple yet strong pipeline for driving topology reasoning. arXiv preprint arXiv:2310.06753  (2023)

\bibitem{ScalableMap}
Yu, J., Zhang, Z., Xia, S., Sang, J.: Scalablemap: Scalable map learning for online long-range vectorized hd map construction. arXiv preprint arXiv:2310.13378  (2023)

\bibitem{hivt}
Zhou, Z., Ye, L., Wang, J., Wu, K., Lu, K.: Hivt: Hierarchical vector transformer for multi-agent motion prediction. In: Proceedings of the IEEE/CVF Conference on Computer Vision and Pattern Recognition. pp. 8823--8833 (2022)

\end{thebibliography}
\end{document}